\documentclass{article}
\usepackage[final]{corl_2020} 
\usepackage{latexsym, multirow,tikz}
\usepackage{amsmath, amsfonts, graphicx, float, wrapfig, tabularx, pgfplots, amssymb}
\usepackage{natbib}
\usepackage{caption} 
\usetikzlibrary{positioning, automata}
\usepackage[ruled,vlined]{algorithm2e}
\usepackage[final]{corl_2020} 
\usepackage{xspace}
\usepackage{multirow}

\usepackage{natbib}
\usepackage{paralist}

\usepackage{array}
\newcolumntype{x}[1]{>{\centering\arraybackslash\hspace{0pt}}p{#1}}

\usepackage{times}

\usepackage{soul}
\usepackage{url}
\usepackage[utf8]{inputenc}
\usepackage{graphicx}
\usepackage{amsmath}
\usepackage{booktabs}
\usepackage{enumitem}
\usepackage{microtype}
\usepackage{pgfplotstable,booktabs}
\usepackage{subcaption}

\newcommand{\eg}{e.g.,\xspace}

\newcommand{\ie}{i.e.,\xspace}

\makeatother

\definecolor{Red}{rgb}{1,0,0}
\definecolor{Green}{rgb}{0,0.69,0}
\definecolor{Blue}{rgb}{0,0,1}
\definecolor{LightBlue}{rgb}{0,0.5,1}
\definecolor{veryLightBlue}{rgb}{0.85,0.98,1}
\definecolor{veryLightGreen}{rgb}{0.6,1,0.6}
\definecolor{Skin}{rgb}{1,0.71,0.69}
\definecolor{Grey}{rgb}{0.5,0.5,0.5}
\definecolor{LightGrey}{rgb}{0.6,0.6,0.6}
\definecolor{Black}{rgb}{0,0,0}
\definecolor{White}{rgb}{1,1,1}
\definecolor{brickred}{rgb}{0.8, 0.25, 0.33}

\usepackage{datatool}

\makeatletter
\providecommand*{\xdtlgetrowindex}[4]{%
  \protected@edef\dtl@dogetrowindex{\noexpand\@dtlgetrowindex{\noexpand#1}{#2}{\number#3}{#4}}%
  \dtl@dogetrowindex
}
\makeatother

\usepackage{mathtools}
\usepackage{cleveref}

\usepackage{scalerel}

\def\ltlX{\mathbin{\scalerel*{\bigcirc}{\forall}\hspace{0.1ex}}}

\title{Learning a natural-language to LTL\\executable semantic parser for grounded robotics}

\author{
  Christopher Wang\\
  \small{MIT CSAIL \& CBMM}\\
  \texttt{czw@mit.edu}\\
  \And
  Candace Ross\\
  \small{MIT CSAIL \& CBMM}\\
  \texttt{ccross@mit.edu}\\
  \And
  Yen-Ling Kuo\\
  \small{MIT CSAIL \& CBMM}\\
  \texttt{ylkuo@mit.edu}\\
  \AND
  Boris Katz\\
  \small{MIT CSAIL \& CBMM}\\
  \texttt{boris@mit.edu}\\
  \And
  Andrei Barbu\\
  \small{MIT CSAIL \& CBMM}\\
  \texttt{abarbu@mit.edu}
}

\begin{document}
\maketitle

\begin{abstract}
  Children acquire their native language with apparent ease by 
  observing how language is used in context and attempting to use it
  themselves.
  They do so without laborious annotations, negative examples, or even
  direct corrections.
  We take a step toward robots that can do the same by training a grounded
  semantic parser, which discovers latent linguistic representations that can be used for the execution of  natural-language commands.
  In particular, we focus on the difficult domain of commands with
  a temporal aspect, whose semantics we capture with 
  Linear Temporal Logic, LTL.
  Our parser is trained with pairs of sentences and executions as well as an
  executor.
  At training time, the parser hypothesizes a meaning representation for the input as a formula in LTL.
  Three competing pressures allow the parser to discover  meaning from
  language.
  First, any hypothesized meaning for a sentence must be permissive enough to reflect all the annotated execution
  trajectories.
  Second, the executor --- a pretrained end-to-end LTL planner --- must find that the observed
  trajectories are  likely executions of the meaning.
  Finally, a generator, which reconstructs the original input, encourages the model to find representations that conserve knowledge about the command.
  Together these ensure that the meaning is neither too general nor too
  specific.
  Our model generalizes well, being able to parse and execute both
  machine-generated and human-generated commands, with near-equal accuracy,
  despite the fact that the human-generated sentences are much more varied and complex with
  an open lexicon.
  The approach presented here is not specific to LTL: it can be applied to any domain
  where sentence meanings can be hypothesized and  an executor can
  verify these meanings, thus opening the door to many applications for robotic agents.
    
\end{abstract}


\vspace{-1ex}
\section{Introduction}
\label{sec:intro}

Natural language has the potential to be the most effective and convenient way to issue commands to a robot. 
However, machine acquisition of language is difficult due to the context- and speaker-specific variations that exist in natural language.
For instance, English usage differs widely throughout the world: between children and
adults, and in businesses vs. in homes.
This does not pose a significant challenge to human listeners because we acquire
language by observing how others use it and then attempting to use it ourselves.
Upon observing the language use of other humans, we discover the latent structure of
the language being spoken.
We develop a similar approach for machines.

Our grounded semantic parser learns the latent structure of natural language utterances.
This knowledge is executable and can be used by a planner to run commands.
The parser only observes how language is used: what was said \ie the command,
and what was done in response \ie a trajectory in the configuration space of
an agent.
We provide two additional resources to the parser, which children also have
access to.
First, we build in an inductive bias for a particular logic, in our case Linear Temporal
Logic (LTL) over finite sequences.
We believe this is a reasonable starting place for our model, since it is widely assumed that priors over
possible languages are built in by evolution and are critical to human language
learning \citep{chomsky2007approaching}.
Second, we provide feedback from an executor: a planner trained end-to-end that is capable of executing formulas
in whatever formalism we use in the prior, in our case LTL.
With only this knowledge, our grounded semantic parser learns to turn sentences
into LTL formulas and to execute those formulas.

The parser strives to find an explanation for what a sentence might mean by
hypothesizing potential meanings and then updating its parameters depending on
how suitable those meanings were.
Four sources of information combine together to inform the semantic parser and
are woven together into a single loss function.
First, all outputs are verified to be syntactically, but not semantically valid
LTL formulas, \ie only valid LTL formulas are accepted.
Second, the parser aims to create interpretations that are generic enough and whose
LTL formulas actually admit the behavior that was observed.
Third, given an interpretation of a sentence, the executor validates that the
observed behavior is rational, \ie has a high likelihood conditioned on that
interpretation.
Fourth, a generator attempts to reconstruct the input to maximize the knowledge
conserved when translating sentences into some formalism.
We do not provide any LTL-specific or domain-specific knowledge.

Our approach performs well on both machine and human generated sentences.
In both cases, it is able to execute around 80\% of commands successfully.
A traditional fully-supervised approach on the machine-generated
sentences outperforms ours with 95\% accuracy, but requires the ground-truth
LTL formulas.
Where this approach of discovering latent structures shines is on real-world
data.
We asked human subjects, unconnected with this research and without knowledge of
robotics or ML, to produce sentences that describe the behavior of robots.
This behavior was produced according to an LTL formula that we randomly generated, but the humans
were free to describe whatever they wanted with whatever words they desired as
long as it was true.
Despite the human sentences being much more varied and complex, including
structures which our formalism cannot exactly represent, our method still finds
whatever latent structure is present and required to execute the
natural-language commands produced by humans with nearly the same accuracy as
those produced by machines.

Executing LTL commands and understanding the kinds of temporal relations that
require LTL is particularly difficult due to the richness and openness of natural language
\cite{brunello2019LTL}.
But LTL is just a stepping stone.
It remains an important open question in grounded robotics: what representation
will be enough to capture the richness of how humans use language?
For instance, notions such as modal operators to reason about hypothetical futures will likely
be required, but what else is unclear.
Having a general-purpose mechanism for learning to execute commands is extremely
helpful under these conditions; we can experiment with different logics and
domains with the same agents by changing the priors and planners while leaving
the rest of the system intact.

\begin{figure}
    \centering
    \includegraphics[width=0.8\linewidth]{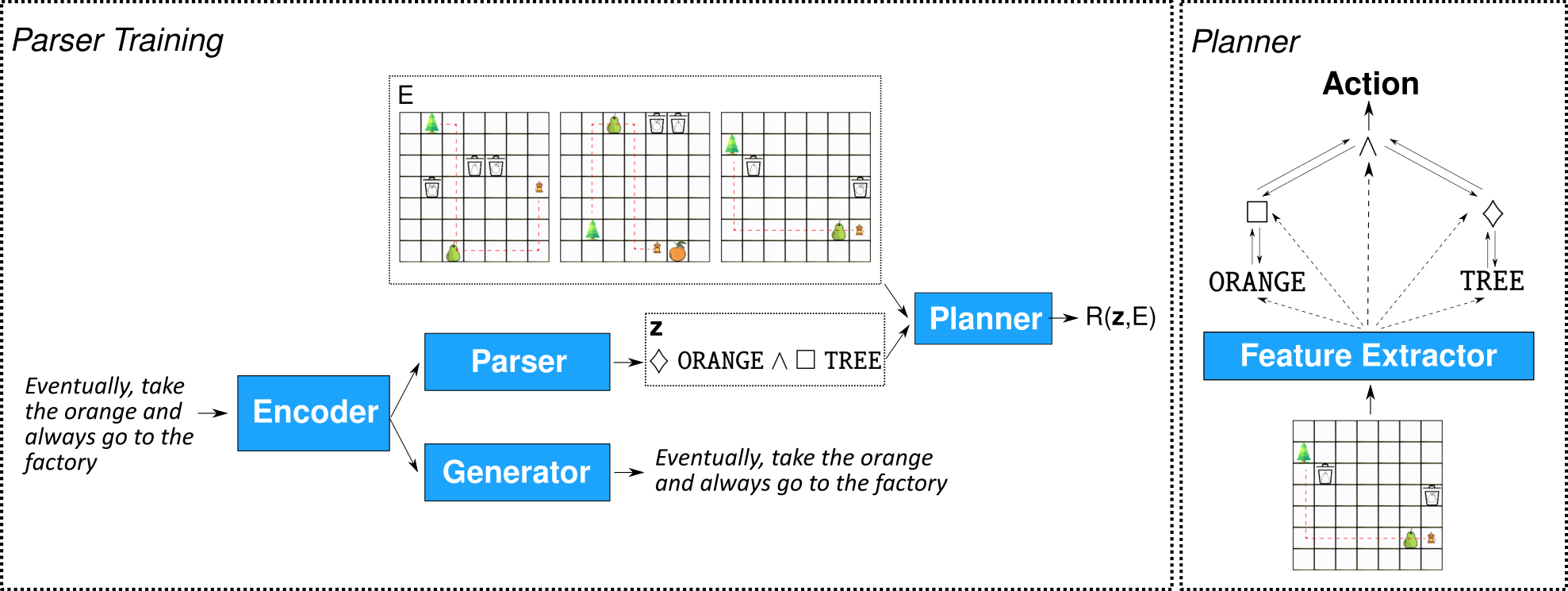}
    \caption{(left) The paradigm for training the parser is shown on the left.
    Given an input sentence, the parser proposes an LTL formula, $\mathbf{z}$, that could encode the meaning of that sentence. 
    That formula is executed by a robotic agent to estimate the likelihood of the observed behavior, $E$, given the interpretation of the sentence, \ie determining whether many more efficient approaches to executing this formula might exist.
    This likelihood is used to compute the reward $R(\mathbf{z},E)$.
    At the same time, a generator attempts to reconstruct the sentence. 
    (right) An example of the planner in action. 
    We use the planner described by \citet{kuo2020encoding} which learns to execute LTL formulas end-to-end from images to actions. 
    Each predicate, \textsc{orange} and \textsc{tree} here, and each operator, are neural networks which together output an action that the robot should take given the current state of the world. 
    Not shown are recurrent connections that enable each component to keep track of execution progress.}
\label{fig:model}
\vspace{-3ex}
\end{figure}

The main contributions of this work are:
\begin{compactenum}
\item a semantic parser that maps natural language to LTL formulas trained without access to any annotated formulas --- no annotations were even collected for human-generated commands,
\item a variant of Craft by \citet{andreas2017modular} suited for grounded
  semantic parsing experiments, and
\item a recipe for creating grounded semantic parsers for new domains that
  results in executable knowledge without annotations for those domains.
\end{compactenum}

\section{Related Work}
\label{sec:related_work}

\paragraph{Semantic parsing}
Early attempts at semantic parsing, such as the famous SHRDLU system  \cite{winograd1971procedures}, were rule-based language systems.
Later, advances in statistical language modeling gave rise to grammar-based approaches that were trained on full supervision \citep{zelle1996learning, zettlemoyerlearning}.
Recent approaches have focused on training grounded semantic parsers using weak supervision.
The work of \citep{berant2013semantic, liang2016neural, liang2018memory, agarwal2019learning} trains a language model on a dataset of question-answer pairs to produce queries in a formal language.
The approaches of \citep{artzi2013weakly, williams2018learning} present a planner and CCG-based parser to generate programs for a deterministic robotic simulator.
However, in these approaches, it is not clear whether constraints on behavior across time are learned, because the tasks do not require this knowledge and the formalisms used cannot easily represent such concepts.
\citet{ross2018grounding} use videos to supervise a grounded semantic parser. 
Although the videos contain information about events across time, the predicate logic used in \cite{ross2018grounding} does not contain the operators necessary to represent the constraints that we focus on in this work.
Nor did that work result in executable knowledge, \ie plans that could drive a robot, merely descriptions of videos and actions.
The semantic parsing literature is relatively sparse on the topic of time and
temporal relations; for example \citet{lee-etal-2014-context} parse
natural-language expressions denoting relative times, in many ways a much easier
task than parsing into LTL.
Even among the parsing approaches that do use the LTL formalism, many require a repository of general-purpose templates \cite{dwyer1999patterns, gruhn2006patterns,konrad2005real,nikora2009automated} and do not truly address unbounded natural language input.

Our work follows recent approaches that cast the problem of semantic parsing as a machine translation task.
Instead of using chart parsers, as is typical for grammar-based approaches \citep{artzi2013weakly, ross2018grounding, williams2018learning}, we use an encoder-decoder sequence-to-sequence model where the input and output are sequences of tokens: in our case, natural language commands and LTL formulas, respectively.
Attempts to train sequence models using weak supervision usually warm start the model by pre-training with full supervision \cite{goldman-etal-2018-weakly, jehl2019learning}.
By contrast, we train our model from scratch using a relatively small set of command-execution pairs. This is difficult in the initial stages of training due to the large exploration space.
To address these challenges, we follow \citet{guu2017language} and \citet{liang2016neural} in using randomized search with a search space restricted to formulas that are syntactically valid in a target logic, LTL.
\begin{figure}
    \centering
    \begin{subfigure}{3cm}
        \includegraphics[width=\linewidth]{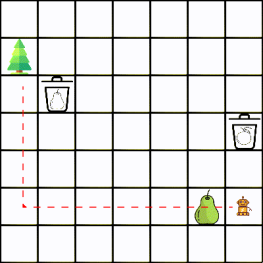}
    \end{subfigure}
    \qquad
    \begin{subfigure}{3cm}
        \includegraphics[width=\linewidth]{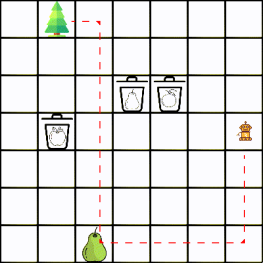}
    \end{subfigure}
    \qquad
    \begin{subfigure}{3cm}
        \includegraphics[width=\linewidth]{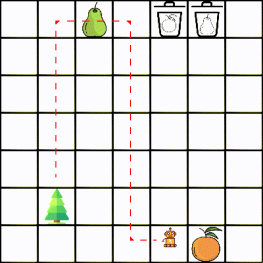}
    \end{subfigure}
    \caption{Execution tracks for the command ``Always take the pear and go to the tree and stay there.''}%
    \label{fig:example}%
\end{figure}

\paragraph{Planning}
Closely related to our task is the work that has been devoted to mapping LTL formulas to execution sequences \citep{sahni2017learning, camancho2019, alshiekh2018safe}.
\citet{kuo2020encoding} presents a compositional neural network that learns to map LTL formulas to action sequences.
It can also compute the likelihood of a sequence of actions conditioned on a formula.
We adopt this agent as the executor in this work and train it for our domain; it never sees a single natural-language utterance, it merely learns how to execute LTL formulas.

Most relevant to our work, \citet{roma2019} trains a weakly supervised semantic
parser for LTL formulas.
This work uses an algorithm that requires significant knowledge
about LTL and is entirely LTL-specific, while here we merely reject
syntactically invalid formulas.
Their approach requires access to the ground truth observations of the
environment to execute LTL formulas step by step while our approach takes as
input images observed by the robot of its environment.
Fundamentally, their approach requires reasoning about locations and paths,
while our approach includes object interactions.
No prior work or prior dataset includes a robot that can interact with objects,
that perceives its environment, and must understand natural-language commands
that have a temporal aspect to them and thus require LTL.

\section{Model}
\label{sec:model}

\Cref{fig:model} (left) provides an overview of our model.
Following one-to-many sequence-to-sequence approaches for multi-task learning such as \cite{dong2015multi}, our model consists of one encoder and two decoders.
A natural language input command, $\mathbf{x}$, is first encoded to a
high-dimensional feature vector then separately decoded into the predicted LTL
formula, $\hat{\mathbf{z}}$, and reconstructed command, $\hat{\mathbf{x}}$.
During training, the planner assigns a reward, $R(\hat{\mathbf{z}}, E)$, to each
hypothesized formula, $\hat{\mathbf{z}}$.
The training dataset, $D=\{(\mathbf{x}, E)\}$, contains pairs
of natural language commands $\mathbf{x}$ and execution demonstrations $E$.
%
Each demonstration, $E$, consists of $k$ trajectory-environment pairs
$E=\{\mathbf{y}_k, e_k\}_{i=1}^{k}$, see \Cref{fig:example}.
A trajectory is a sequence of actions \eg
$\textbf{y} = [\texttt{left}, \texttt{left}, \texttt{up}, \texttt{grab}]$.

\subsection{Parser and Generator}
\label{sec:parser}

Our architecture is similar to the sequence-to-sequence models of \citet{guu2017language} and \citet{dong2016language}.
Given input sentence $\mathbf{x}$, our model defines a probability for an output formula $\mathbf{z} = [z_1,...,z_m]$:
\begin{equation}
p
\left(
\mathbf{z} \mid \mathbf{x}
\right)
=
\prod_{i=1}^{|\mathbf{z}|}
p
\left(
z_i \mid z_{<i}, \mathbf{x}
;
\theta_{\text{parse}}
\right)
\label{model_prob}
\end{equation}
where $\theta_{\text{parse}}$ are the parser parameters.

\paragraph{Encoder}
The encoder is a stacked bi-directional LSTM that takes word embeddings as
produced by the pretrained English GloVe model \cite{pennington2014glove}.
The encoder maps the input $[x_1,...,x_n]$ to a feature representation
$\mathbf{h} = [h_1,...,h_n]$.

\paragraph{Parser-Decoder}
The parser decoder takes the feature vector from the encoder and generates a
sequence of tokens: the LTL formula $\mathbf{\hat{z}} = [z_1,...,z_m]$.
The decoder is a stacked LSTM with an attention mechanism
\cite{bahdanau2015attention}, as implemented by \citet{bastings2018annotated}.
Dropout is applied before the final softmax.

At each step, the attention mechanism produces a context vector $c_i$ from the decoder hidden state $s_i$ and  $\mathbf{h}$. 
The previous decoder input $z_{i-1}$, along with $c_i$ and $s_i$, are used to produce a distribution over output tokens:
\begin{equation}
p(z_i \mid z_{<i}, \mathbf{x}; \theta_{\text{parse}}) = \text{Softmax} (W_o[z_{i-1}; c_i; h_{i}])
\end{equation}
We keep a stack of generated tokens and output LTL formulas in post-order, since all operators in LTL have a fixed arity.
This allows us to avoid the problem of parentheses matching.
We condition the decoder to sample syntactically-valid continuations of the
formula being decoded.
Note that this assumes nothing about the formula's meaning.
We simply build in the fact that certain
continuations are trivially guaranteed to never be syntactically valid; for
example, the formula $\wedge\wedge$ is not a valid LTL formula.
Practically, this property is trivially computable since LTL formulas, and
virtually all logics in general, use notation that is context-free, which allows us
to remove invalid options from the softmax output before sampling the next
token.
Following \citet{guu2017language}, we use $\epsilon$-randomized sample decoding for better exploration during training.
At each timestep $i$, with probability $p=\epsilon$, we draw the next token $z_i$ according to $p(z_i \mid z_{<i}, \mathbf{x}; \theta_{\text{parse}})$.
With probability $p=1-\epsilon$, we pick a valid continuation uniformly at random.

\paragraph{Generator-Decoder}

The parser-decoder produces formulas which are scored by the planner, but this
does not ensure that the full content of the utterance is reflected in the
parse.
To encourage this, we include a second, separate decoder, which is trained to reconstruct the
original natural language command, $\mathbf{x}$, from the feature representation
$[h_1,...,h_n]$.
This is a standard multi-task learning approach that is often used in the
literature to improve generalization.
The architecture of this component is identical to the parser-decoder.

\begin{table*}
\begin{tabular}{@{}l@{\hspace{0.7ex}}l@{\hspace{0.7ex}}l}
   & \emph{Constituents} &  \emph{Description} \\ 
 \emph{Logical operators} & $\wedge$, $\vee$ &  And, or \\ 
 \emph{Temporal operators}  & $\square$, $\Diamond$, \textsf{U} & Eventually, always, until\\ 
 \emph{Objects} & \textsc{Apple}, \textsc{Orange}, \textsc{Pear} & Objects that can be held\\ 
 \emph{Relations} & \textsc{Closer\_apple}, \textsc{Closer\_orange}, \textsc{Closer\_pear} & Spatial relations\\ 
 \emph{Destinations} & \textsc{Flag}, \textsc{House}, \textsc{Tree} & Destinations \\ 
\end{tabular} 
\label{vocab}
\caption{The various constituents of our target formalism are shown above. We ground the
  meaning of natural-language sentences produced by humans into LTL$_f$
  (\citep{de2013linear}) without negation. Predicates from
  the Craft domain are renamed as users found these labels easier to
  understand. Note that this is the size of the target formalism; it is
  unrelated to the complexity of the input. Humans produced sentences that
  contained 266 words across them which had to be grounded to these semantics.}
\end{table*}

\paragraph{Planner}

Formulas are scored using a planner.
We adopt the one described by \citet{kuo2020encoding} because it learns to
execute LTL formulas end to end and is pretrained for the \textsc{Craft} environment
that we evaluate on.
Given a formula and an environment, the planner is trained to create an execution sequence; it never has
access to the natural language utterance.
It learns to extract features from images of the environment around the robot and
acquires knowledge about LTL predicates and operators in order to execute novel formulas in
novel environments.
\Cref{fig:model} (right) shows an example of the planner configured to execute
an LTL formula.
Given an LTL formula, the planner is configured by assembling a
compositional recurrent network specific to that formula; it then guides the
robot to execute the formula.
This compositionality enables zero-shot generalization to new formulas.
Any planner could in principle be used as long as it could learn to execute
formulas for the target domain and if it could score an arbitrary trajectory
against a formula.

The planner does not have access to the ground truth environment.
Instead, it observes an image of a $5\times 5$ patch of the world around it, which
is passed through a learned feature extractor CNN.
At every time step, each module within the planner takes as input the features
extracted from the surroundings of the robot, the previous state of that module,
and the previous state of the parent.
The state of the root of the LTL formula, according to an arbitrary but consistent
parse of the formula, is decoded to predict a distribution over actions.
The model is pretrained on randomly generated environments and LTL formulas using
A2C, an Advantage Actor-Critic algorithm \citep{sutton2000policy,
  mnih2016asynchronous}.

\section{Training}
\label{sec:training}

The model described above produces a candidate LTL formula
$\mathbf{\hat{z}}$, along with a reconstruction of the input, $\mathbf{\hat{x}}$.
Each candidate formula is used to compute a reward that incorporates the
likelihood, as
computed by the planner, of the observed trajectories given the hypothesized LTL formula.
This plays two roles: first it ensures that the observed trajectories are
actually feasible given the hypothesized LTL formula; otherwise they will have
zero likelihood. Secondly, it provides a score for how rational the planner judges the behavior to be.
Not all feasible paths are equally rational, and so by extension not equally
likely.
For example, a complex observed trajectory is unlikely to be the consequence of a simple command:
it is more likely that the parser is producing an overly broad interpretation rather
than the observed trajectory going out of its way to do something unnecessary.
The reward is then
\begin{equation}
R(\hat{\mathbf{z}}, E)
=
\begin{cases}
    \frac{1}{k}
    \sum_{(e_i, \mathbf{y}_i)}
    \frac{1}{|\mathbf{y}_i|}
    \sum_{j} 
    p(\mathbf{y}_{i,j}
    \mid \mathbf{\hat{z}},
         \mathbf{y}_{i,<j}, e 
                
    ) & \forall \mathbf{y} \in Y. \mathbf{y} \in \hat{\mathbf{a}} \\
    0 & \text{otherwise}
\end{cases}
\end{equation}
where $\mathbf{\hat{a}}$ is an NFA representation of the formula $\hat{\mathbf{z}}$, so that $\mathbf{y} \in \hat{\mathbf{a}}$ indicates that $\mathbf{y}$ is feasible.
We optimize the reward of the output with either REINFORCE
\cite{williams1992simple} or Iterative Maximum Likelihood (IML).
This reward computes an average of the likelihood over the $k$ execution traces in $E$, conditioned on the candidate
formula $\mathbf{\hat{z}}$ (recall that each sentence is paired with $k$
demonstration trajectories, each in a different randomly-generated environment).
%
%
Since the size of the search space for LTL formulas grows exponentially in the
length of the formula, we employ curriculum learning as in
\citet{liang2016neural}.
Every 10 epochs, we increase the maximum length of the predicted formulas by 3.

\subsection{Reinforcement Learning}

In the reinforcement learning setting, our objective is to maximize the expected reward, marginalizing over the space of possible formulas:
$
J_{RL} =
\sum_{\mathbf{x}}
\sum_{\mathbf{z}}
R(\mathbf{z})
p_{\theta}
\left(
\mathbf{\hat{z}} \mid \mathbf{x}
\right)
\label{expected_reward}
$.
%
%
We use the REINFORCE algorithm \cite{williams1992simple} to learn the policy
parameters with Monte-Carlo sampling.
For better exploration, we use $\epsilon$-dithering when sampling as described
in \ref{sec:parser}.
To incorporate the generator, we optimize a linear combination of this reward and
the reconstruction loss,
$J_{GEN}=-\sum_{\mathbf{x}}\log p(\mathbf{x} \mid \mathbf{x}; \theta_{GEN})$,
so that $J = J_{RL} + \alpha J_{GEN}$.
We adjust $\alpha$ at training time to balance the two components.
In
particular, it is important to start with a small $\alpha$ initially, since $J_{RL}$
is small when $\theta$ is untrained and few candidate formulas have non-zero reward.


\subsection{Iterative Maximum Likelihood}

Iterative Maximum Likelihood, IML has proven itself to be as efficient if not
more efficient when acquiring semantic parsers compared to RL.
We adopt a method similar to that of \citep{liang2016neural} and
\citep{agarwal2019learning}.
First, we explore the output space by sampling $K$ formulas from the parser.
We keep the highest reward formulas $\hat{z}^*$ and use them as a pseudo-gold.
We then maximize the likelihood of the pseudo-gold formulas over the course of
10 epochs:
$ J_{IML} = \sum_\mathbf{x} \sum_{i} \log p(\hat{\mathbf{z}}_i^* \mid
\mathbf{x}; \theta_{\text{parse}}) $.
To incorporate the generator, we again combine the two objective
functions, but this time no scaling parameter is required:
$J = J_{IML} + J_{GEN}$.
Sampling and MLE steps are then iterated.

%

\section{Experiments}

We test the parser in two experiments.
The first verifies that machine-generated natural-language commands derived
from LTL formulas can be understood and followed by a trained agent in a way that reflects the formula correctly.
The second verifies that our model can 1) understand sentences produced by humans which describe a given behavior
and 2) express a plan in LTL that 
will result in this behavior.
Note that the humans never see the LTL formulas; they produce natural language
descriptions for the behavior of robots.

In all cases, our model has the same hyperparameters.
The stacked LSTMs all have two layers with hidden dimensions of size 1000 and
dropout probability $0.2$.
We use Adam with learning rate $1e-3$.
REINFORCE and IML both sample $128$ formulas to compute the expectation and
generate sentences for the next iteration with $\epsilon=0.15$ when exploring.
We train using $k=3$ trajectories for 50 epochs.
Results are reported for the model with highest validation set performance as
measured by the \textbf{Exec} metric, see \cref{sec:results}.
At test time, we decode using a beam search with width 10.

\paragraph{Temporal Phenomena}

Most randomly generated LTL formulas are uninteresting, similar to the way in which most instances
of the boolean satisfiability problem SAT are uninteresting \citep{horie1997hard}.
To avoid such issues, we adopt the standard classification of LTL formulas
produced by \citet{Manna:89} and generate formulas in their six
partially-overlapping categories:
%
%
\emph{safety}, \emph{guarantee}, \emph{persistence}, \emph{recurrence}, \emph{obligation}, and \emph{reactivity}. 
Respectively, \emph{safety}, \emph{guarantee}, \emph{persistence}, and
\emph{recurrence}, ensure that a property will always hold, will hold at least
once, will always hold after a certain point, or will hold at repeated points
in time.
\emph{Obligation} and \emph{reactivity} are compound classes formed by
unrestricted boolean combinations of the safety and recurrence classes
respectively.
%
%
While we adopt the LTL over finite sequences, LTL$_f$ as described by
\citet{de2013linear}, the target formalism uses the $\Diamond$ \emph{eventually}
and $\square$ \emph{always} temporal operators rather than $\ltlX$ \emph{next}
to readily generate instances of the \citet{Manna:89} classes.
Since we found humans to be very unlikely to spontaneously generate sentences
that required negation, it was not included.
The components of the target formalism are shown in \cref{vocab}.


\begin{wraptable}[17]{R}{0.4\textwidth}
\small
\vspace{-2ex}
\begin{tabularx}{\linewidth}{Xl}\\
\# Total sentences & 2,000 \\  \midrule
\# \textbf{Machine sentences} & 1,000 \\
\# Guarantee & 204 \\ 
\# Safety & 264 \\
\# Recurrence & 243 \\
\# Persistence & 214\\
\# Obligation & 52\\
\# Reactivity & 23\\
Avg. words/sent. & 17.7 $\pm$ 8.4 \\
\# Lexicon size & 44 \\\midrule
\# \textbf{Human sentences} & 1,000 \\ 
Avg. words/formula & 5.2 $\pm$ 2.9 \\
Avg. words/sent. & 8.3 $\pm$ 3.3 \\
\# Lexicon size & 266 \\
\end{tabularx}
\caption{Dataset statistics. Note that the human-generated data is far more
  varied with a much larger lexicon.}\label{wrap-tab:1}
\end{wraptable} 

\paragraph{Mechanical dataset}

We collect two sets of data.
The first, a mechanically-generated dataset, consists of 1,000 natural language sentences paired with 3,000 execution traces, 3 per example, with a 70/15/15 training/val/test split.
All commands and environments are given in the context of the Minecraft-like \textsc{Craft} environment, which we adapt from  \citet{andreas2017modular} and \citet{kuo2020encoding}.

Following \citet{jia2016recombindation} and \citet{goldman-etal-2018-weakly}, we
generate sentences and formulas by randomly and uniformly sampling productions
and terminals from a synchronous context-free grammar (Appendix \ref{sec:grammar}).


For each command-formula pair, we populate three 7x7 grid environments with
objects and landmarks.
Each environment includes all the items and landmarks in the corresponding
command in random locations.
Other objects and landmarks that are not in the command are each included with
probability $0.3$, resulting in somewhat densely populated environments.
Of course, this does not guarantee that a command can be executed on a
particular map.

Given the formula, we generate a non-deterministic finite state automaton using
Spot \cite{duret2016spot}.
An oracle brute-force searches the action space and generates an action sequence
which the automaton accepts; this can be quite slow.
Rejection sampling over this process results in three environments for each command-formula pair.
LTL with finite semantics \citep{dutta2014assertion,de2013linear} requires a
time horizon: we set it at 20 steps, by which point the command must be satisfied,
or equivalently, the automaton must be in an accepting state.

Some commands mandate that a condition hold globally, \eg ``Always hold the gem''. 
However, unless the agent's initial state satisfies this condition, \eg the
robot happens to start with the gem in hand, no satisfying action sequence is
possible.
To address this, we allow the robot time to approach and grab the object, which is
surely the intent of any human speaker, replacing each predicate $p$ with
$\texttt{closer}(p) \textsf{ U } p$, where $\texttt{closer}(p)$ means ``closer
to $p$''.
That is, instead of requiring that $p$ be satisfied immediately and always, we mandate that the robot move
closer to $p$ until $p$ is satisfied.

\paragraph{Human-generated dataset}

We take all sampled environments from the mechanical dataset and present them to
humans.
Note that humans only see what the robots do, not why they did it.
They do not see the LTL formulas or the machine-generated utterances.
We asked six human annotators, who were working for pay, unconnected to this research,
and unfamiliar with NLP or robotics,
to describe what the robots are doing.
Of course, this leads to different sentences than those that originally
generated the behavior of the robots.

This process ensures that even though our mechanical dataset was generated from
a context free grammar, no trace of that grammar remains; humans generate the
sentences they are comfortable with.
The distractor objects and landmarks were not removed for this experiment,
giving annotators the opportunity to refer to them.
The target object and the intended actions need never appear in the final human
descriptions.
We did not collect LTL formulas from the human annotators.
Note that the size of the lexicon, 266 words, that the humans used is far larger than both what our formalism
supports and what was produced by the machines.

\subsection{Results}
\label{sec:results}

Results are shown in \cref{tab:results}.
The machine-generated data is annotated with ground truth LTL formulas, but no equivalent
concept exists for the human-generated dataset, since the humans only had to explain what
they thought the robots were doing; it might not even be possible to fully capture
the semantics of their sentences by LTL.
\textbf{Exec} measures the fraction of formulas that accept all $k$ execution
traces.
This is an overestimate of the performance of the grounded parser; merely
accepting formulas does not guarantee any understanding.
\textbf{Plan} measures the fraction of the environments that the planner
executes correctly on, given the predicted formula as input.
This is an underestimate of the performance of the grounded parser; even a
human controlling a robot in such environments may not exactly carry out the
expected actions, since many formulas are hard to interpret and there is much opportunity for error.
As is common in linguistics, no single metric perfectly captures performance, but these two
metrics do bracket the performance of the approach.

\begin{table}[t]
    \begin{minipage}[t]{.5\linewidth}
    \centering
    Machine-generated dataset\\[2ex]
    \begin{tabular}{lcccc|}
               & \textbf{Exec} &\textbf{Plan} & \textbf{Seq} & \textbf{Exact} \\ 
      Supervised   & 94.7 & 36.7 & 94.9 & 91.3 \\  \midrule
      RL           & 82.0 & 41.3 & 22.9 & 8.7 \\ 
      RL + generator & 83.3 & 41.3 & 23.9 & 8.7 \\ 
      IML          & 81.3 & 32.2 & 14.0 & 2.0 \\ 
      IML + generator & 85.3 & 34.9 & 15.0 & 4.0 \\ 
    \end{tabular} 
    \end{minipage}
    \begin{minipage}[t]{.4\linewidth}
    \centering
    Human-generated dataset\\[2ex]
    \begin{tabular}{@{}lcc}
               & \textbf{Exec} &\textbf{Plan}  \\
      Random       & 16.3 & 17.5 \\ \midrule
      RL           & 78.7 & 40.7 \\ 
      RL + generator & 79.3 & 43.3 \\ 
      IML          & 80.0 & 28.7 \\ 
      IML + generator & 83.3 & 31.8 \\ 
    \end{tabular} 
    \end{minipage}
    \caption{Results on the machine-generated (left) and the human-generated
      (right) datasets. \textbf{Exec} measures the likelihood of formulas that
      recognize the ground truth trajectories, an overestimate of the real
      performance of the parser. \textbf{Plan} measures how often the planner
      produced a correct trajectory given the predicted formula, an
      underestimate of the real performance. \textbf{Seq} and \textbf{Exact}
      measure the overlap between predicted and ground-truth LTL formulas; note
      that many formulas have identical semantics, even humans may do poorly on
      this metric. The fully-supervised method outperforms our approach, as
      expected, but it is only relevant for the machine dataset where
      ground-truth annotations exist. Note that the drop between the machine-
      and human-generated datasets is small, despite the human sentences being
      more diverse.}
    \label{tab:results}
    \vspace{-5ex}
\end{table}

A more stringent metric would be to investigate how the annotated and predicted
LTL formulas compare, which is possible on the machine-generated dataset.
\textbf{Exact} measures the fraction of predicted formulas which are equivalent to the
ground truth.
\textbf{Seq} is the F1 token overlap score between the predicted formula and the
ground truth.
These are extremely stringent metrics that even humans would perform poorly on, as the
same actions can be carried out for many different reasons and many LTL formulas
are equivalent in context.
Typical mistakes made when predicting the exact formula on the machine-generated
dataset are shown on the right in \cref{tab:example_out}.

\begin{wraptable}[13]{R}{0.4\textwidth}
\vspace{-2.5ex}
\small
\begin{tabularx}{\linewidth}{l@{\hspace{1ex}}X}
 Input & \textit{Either grab the apple or the pear and hold them forever.} \\[1ex]
 Target & $\square \Diamond \textsc{Flag} \wedge \square \Diamond \textsc{Orange}$ \\ 
 RL+gen & $\square \Diamond (\square \Diamond \textsc{Flag} \vee \Diamond \textsc{Orange})$ \\ 
 IML+gen & $\square \Diamond (\textsc{Flag} \vee \textsc{Orange})$ \\
\end{tabularx}
\vspace{-1ex}
\caption{Predicted output for the machine-generated test set showing typical mistakes. 
  These formulas are hard to tell apart from observations of the
  robot's behavior. 
  This makes it harder to learn the correct form while at the
  same time encouraging correct executions (See Appendices \ref{sec:sample_output} and \ref{sec:qualitative}).
  }
\label{tab:example_out}
\end{wraptable} 

Overall, the supervised method outperforms our method, but when considering the
percentage of correctly executed formulas, it only outperforms the
weakly-supervised approach by 5-10\%.
Both RL and IML performed well, with the generator increasing performance by
1-4\%.
Overall, IML with the generator was the highest performing approach and recovers
almost all the performance of the supervised approach.

We investigated how performance varied as we provided more examples per sentence
using the machine-generated data.
While the fraction of correctly executed sentences stays roughly the same, the
exact match goes up significantly from 2\% at $k=3$ to $14\%$ at $k=7$.
The ambiguity in this domain prevents exact matches, but allows for good
executions.

On the human-generated dataset, despite the fact that the formalism we use is
small compared to the 266 words that humans used, the grounded parser is able to
understand most commands.
It correctly executes in about 43\% of the environments and accepts about 80\% of
commands.

\section{Conclusion}
\label{sec:conclusion}

We created a grounded semantic parser that, given only minimal knowledge about
its environment and formalism, was able to discover the structure of an input
language and produce executable formulas to command a robot.
Its performance is competitive with a state of the art supervised approach, even
though we provide no direct supervision.
We were able to get similar performance on a challenging dataset produced by
humans that could use any word and sentence construction to describe the actions
of robots, even those that our formalism cannot completely capture.
This model has virtually no knowledge of its domain or target logical formalism;
it merely requires a planner and a method to reject syntactically invalid
formulas.
Many problems in robotics and NLP could be tackled by such an approach because
of its low requirements for annotations.
For example, data already exists to guide agents to reproduce the actions of
customer service agents in response to queries.
And in the robotic domain, future work might involve observing what humans say to one another and
then acquiring a domain-specific semantic parser to guide robots on a worksite for
example.
Being able to adapt to variations in language use and to changes in the
environment is crucial to building useful robots, because the same
language may carry very different meanings in different contexts.
In the long term, we hope that this line of research leads both to robots that
understand us and to robotic systems that can be used to probe how children
acquire language, bringing robotics and linguistics closer together.

%



\clearpage
\acknowledgments{This work was supported by the Center for Brains, Minds and Machines, NSF STC award 1231216, the Toyota Research Institute, the MIT CSAIL Systems that Learn Initiative, the DARPA GAILA program, the United States Air Force Research Laboratory under Cooperative Agreement Number FA8750-19-2-1000, and the Office of Naval Research under Award Number N00014-20-1-2589 and Award Number N00014-20-1-2643. The views and conclusions contained in this document are those of the authors and should not be interpreted as representing the official policies, either expressed or implied, of the U.S. Government. The U.S. Government is authorized to reproduce and distribute reprints for Government purposes notwithstanding any copyright notation herein.}

\bibliography{ref}  

\appendix
\section{Grammar}
\label{sec:grammar}
The grammar used to produce our machine generated commands is shown below:
\\
\begin{align*}
\textsc{BinOp} \rightarrow     & \text{ `and'} \mid \text{`or'}\\
\textsc{UOp} \rightarrow       & \text{ `do not'} \mid \text{`you should not'}\\
\textsc{Item} \rightarrow      & \text{ `apple'} \mid \text{`orange'} \mid \text{`pear'} \\
\textsc{Landmark}  \rightarrow &  \text{ `flag'} \mid \text{`house'} \mid \text{`tree'}\\
\textsc{Predicate} \rightarrow & \text{ `be around the' } \textsc{Landmark} \mid \text{`be near the' } \textsc{Landmark} \\
                               & \mid \text{`go to the' } \textsc{Landmark} \mid  \text{`hold the' } \text{Item}\\
                               & \mid \text{`take the' } \textsc{Item} \mid \text{`possess the' } \textsc{Item}\\
\textsc{P } \rightarrow        & \textsc{ Predicate } \mid \textsc{UOp } \textsc{Predicate } \mid \textsc{Predicate } \textsc{BinOp } \textsc{Predicate } \mid \textsc{UOp } \textsc{P }\\
\textsc{S } \rightarrow        & \textsc{ Safety } \mid \textsc{Guarantee } \mid \textsc{Obligation } \mid \textsc{Recurrence } \mid\\
                               & \mid \textsc{Persistence } \mid \textsc{Reactivity }\\
\textsc{SPrefix } \rightarrow  & \text{ `always'} \mid \text{`at all times,'}\\
\textsc{SSuffix } \rightarrow  & \text{ `forever'} \mid \text{`at all times'} \mid \text{`all the time'}\\
\textsc{Safety } \rightarrow   & \textsc{ SPrefix } \textsc{P } \mid \textsc{P } \textsc{SSuffix } \mid \textsc{Safety } \textsc{BinOp } \textsc{Safety }\\
\textsc{GPrefix } \rightarrow      & \text{ `eventually'} \mid \text{`at some point'}\\
\textsc{NotPredicate } \rightarrow & \textsc{ UOp } \textsc{Predicate }\\
\textsc{Guarantee } \rightarrow    & \textsc{ GPrefix } \textsc{P } \mid \text{`guarantee that you will' } \textsc{Predicate } \\
                                   & \mid \text{ `guarantee that you'} \textsc{NotPredicate } \mid \textsc{Guarantee } \textsc{BinOp } \textsc{ Guarantee }\\
\textsc{Obligation } \rightarrow   & \textsc{ Safety } \textsc{BinOp } \textsc{Guarantee } \mid \textsc{Obligation } \textsc{BinOp } \textsc{Safety }\\
                                   & \mid \textsc{ Obligation } \textsc{BinOp } \textsc{Guarantee }\\
\textsc{Recurrence } \rightarrow   & \text{ `eventually,' } \textsc{P } \text{`and do this repeatedly'} \mid \textsc{Recurrence } \textsc{BinOp } \textsc{Recurrence }\\
\textsc{Persistence } \rightarrow  & \text{ `at some point, start to' } \textsc{P } \text{`and keep doing it'} \\
                                   &\mid \textsc{Persistence } \textsc{BinOp } \textsc{Persistence }\\
\textsc{Reactivity } \rightarrow   & \textsc{ Recurrence } \textsc{BinOp } \textsc{Persistence } \mid \textsc{Reactivity } \textsc{BinOp } \textsc{Recurrence } \\
                                   & \mid \textsc{ Reactivity } \textsc{BinOp } \textsc{Persistence }\\
\end{align*}

\newpage
\section{Sample output predictions}
\label{sec:sample_output}
\subsection{Grammar-generated sentences}
\label{sec:app_examples}
\begin{table}[h]
\caption[Predicted output on the grammar-generated sentences]{Predicted output on the best model parameters for the test set of the grammar-generated commands}
\begin{tabularx}{\linewidth}{lX}
\toprule 
Input & \textit{eventually, be around the tree  or go to the flag and do this repeatedly}\\
Target & $( \square \lozenge ( ( \textsc{ Tree } ) \vee ( \textsc{ Flag } ) ) )$\\
RL+gen & $( \square \lozenge ( ( \textsc{ Flag } ) \vee ( \textsc{ Tree } ) ) )$\\
IML+gen & $( \square \lozenge ( ( \lozenge ( \textsc{ Tree } ) ) \vee ( \square \lozenge ( ( \square ( \textsc{ Tree } ) ) \vee ( \square \lozenge ( \textsc{ Flag } ) ) ) ) ) )$\\
\midrule
Input & \textit{at some point, start to be around the house and go to the flag and keep doing it}\\
Target & $( \lozenge \square ( ( \textsc{ House } ) \wedge ( \textsc{ Flag } ) ) )$\\
RL+gen & $( \square \lozenge ( ( \textsc{ House } ) \vee ( \textsc{ Flag } ) ) )$\\
IML+gen & $( \lozenge \square ( ( \lozenge ( \textsc{ House } ) ) \vee ( \square ( \textsc{ Flag } ) ) ) )$\\
\midrule
Input & \textit{at all times, possess the pear or be around the house and guarantee that you will go to the tree}\\
Target & $( ( \square ( ( \textsc{ Pear } ) \vee ( \textsc{ House } ) ) ) \wedge ( \lozenge ( \textsc{ Tree } ) ) )$\\
RL+gen & $( \square \lozenge ( ( \textsc{ House } ) \vee ( \textsc{ Tree } ) ) )$\\
IML+gen & $( \square \lozenge ( ( \square \lozenge ( \textsc{ Pear } ) ) \vee ( \square \lozenge ( ( \square \lozenge ( \textsc{ Pear } ) ) \vee ( \square ( \textsc{ Pear } ) ) ) ) ) )$\\
\midrule
Input & \textit{at some point, start to be around the house and possess the orange and keep doing it}\\
Target & $( \lozenge \square ( ( \textsc{ House } ) \wedge ( \textsc{ Orange } ) ) )$\\
RL+gen & $( \square \lozenge ( \textsc{ House } ) )$\\
IML+gen & $( \square \lozenge ( ( \square \lozenge ( \textsc{ House } ) ) \vee ( \square \lozenge ( ( \lozenge ( \textsc{ House } ) ) \vee ( \square \lozenge ( \textsc{ House } ) ) ) ) ) )$\\
\bottomrule 
\end{tabularx} 
\end{table}

\subsection{Human-generated sentences}
\begin{table}[h]
\caption[Predicted output on the human-generated sentences]{Predicted output on the best model parameters for the test set of the human-generated sentences}
\begin{tabularx}{\linewidth}{lX}
\toprule 
Input : \textit{get your hands on the orange or the apple some time.}\\
RL+gen: $( \square \lozenge ( \textsc{ Orange } ) )$\\
IML+gen: $( \square \lozenge ( ( \lozenge ( ( \lozenge \square ( \textsc{ Apple } ) ) \vee ( \lozenge ( \textsc{ Orange } ) ) ) ) \vee ( \square \lozenge ( \textsc{ Tree } ) ) ) )$\\
\midrule
Input: \textit{guarantee that you snatch the pear }\\
RL+gen: $( \square \lozenge ( \textsc{ Pear } ) )$\\
IML+gen: $( \square \lozenge ( ( \square \lozenge ( ( \lozenge ( \textsc{ Apple } ) ) \vee ( \square ( \textsc{ Tree } ) ) ) ) \vee ( \square \lozenge ( \textsc{ Pear } ) ) ) )$\\
\midrule
Input: \textit{guarantee that you approach the trash can and visit the tree }\\
RL+gen: $( \square \lozenge ( \textsc{ Tree } ) )$\\
IML+gen: $( \square \lozenge ( ( \square \lozenge ( ( \lozenge ( \textsc{ Tree } ) ) \vee ( \lozenge ( \textsc{ House } ) ) ) ) \vee ( \square \lozenge ( \textsc{ Tree } ) ) ) )$\\
\midrule
Input: \textit{ensure that you pick up the peach and then take it to the trash can}\\
RL+gen: $( \square \lozenge ( \textsc{ Orange } ) )$\\
IML+gen: $( \square \lozenge ( ( \lozenge ( ( \square \lozenge ( \textsc{ Orange } ) ) \wedge ( \square ( \textsc{ Orange } ) ) ) ) \vee ( \square ( \textsc{ Orange } ) ) ) )$\\
\bottomrule 
\end{tabularx} 
\end{table}

\section{Qualitative failure analysis}
\label{sec:qualitative}
\begin{wraptable}[11]{R}{0.4\textwidth}
    \vspace{-7pt}
    \centering
    \begin{tabular}{lr}
                       &\textbf{Avg. Reward}\\
\midrule
RL + gen     & 0.392 \\
IML + gen    & 0.373 \\
Ground truth & 0.369\\
    \end{tabular} 
    \caption[The average rewards received by the predicted formulas]{The average rewards received by the predicted formulas for the best performing parameters on the machine-generated data.}
    \label{tab:reward}
\end{wraptable}

From observing our predicted output, we see the limitations of our approach (\Cref{sec:sample_output}).
While it is encouraging to see that the model generally learns to produce the correct predicates, there are some situational ambiguities that are hard to overcome by observing execution demonstrations.
Namely, it is difficult to distinguish \textit{eventually} $\lozenge$ and \textit{always} $\square$ by observing whether a formula accepts the execution trajectories.
For example, consider the input command \textit{always go to the tree} and a corresponding execution trace $\mathbf{y}$.
Then, a hypothesis such as $\lozenge \textsc{Tree}$ \ie \textit{eventually go to the tree}, will receive a non-zero reward, because it is able to accept the demonstration $\mathbf{y}$.
This is because a trace that shows the robot \textit{always} going to the tree must necessarily show the robot \textit{eventually} going to the tree.
In general, the model can always receive a non-zero reward for proposing $\lozenge$ in a situation which calls for $\square$.
In the same way, any commands involving disjunction are difficult to learn.
That is, the model can always receive a non-zero reward by proposing $p \vee q$ in a situation which requires $p \wedge q$.

These ambiguities would not hinder training if the planner always gave a higher likelihood to the correct formula.
But we find that this is not the case. 
\Cref{tab:reward} shows that on average, the ground truth formula receives a lower reward than the predicted formula,
which indicates that further efforts to maximize the score will not result in more accurate parses.
This problem is likely to become less severe if future advances produce more accurate planners that place higher likelihood on paths which more efficiently execute a formula.

\section{Comparison to baselines}
\label{sec:baseline}
\begin{figure}[h]
    \centering
    \includegraphics[width=0.8\linewidth]{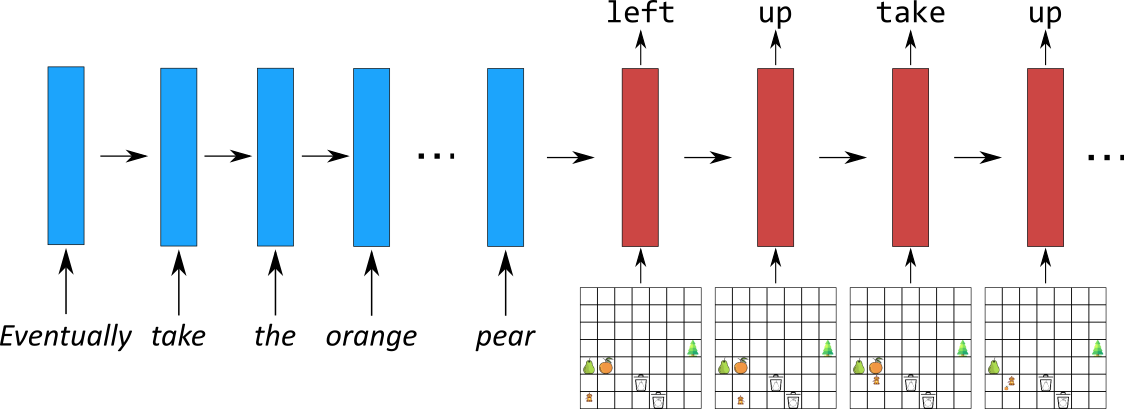}
    \caption{The baseline sequence-to-sequence architecture which encodes natural language into a high dimensional embedding and then decodes the embedding into a distribution over actions. The recurrent units that serve as the encoder and decoder are both stacked ($n=2$) LSTMs with hidden-size of 1000. A feature representation of the environment is extracted using a linear layer.}
    \label{fig:baseline}
\end{figure}
To determine the effectiveness of LTL as an intermediate representation, we compare to a simple end-to-end language-to-action neural network model.
We adapt the architecture from \cite{anderson2018vision} as a baseline for comparison (\Cref{fig:baseline}). 
During training, the model takes a natural language command as input and generates a sequence of actions. 
The loss is simply the likelihood of the ground truth action sequence.
We train on the same set of human- and machine-generated sentences as described in \Cref{sec:training}. 
Teacher-forcing is used when decoding the sequences.

This baseline produces acceptable actions 22\% of the time on our human-generated test set and 18.7\% of the time on machine-generated test set. 
These measures fall below our model's performance of 41.3\% and 43.3\% respectively, as seen in the \textbf{Plan} columns of \Cref{tab:results}. 

A comparison to a more sophisticated language-to-action planner, a variant of which serves as our executor, can be obtained from \cite{kuo2020encoding}, which finds the correct execution 50\% of the time on the test set in a related, but not identical, environment. 

Outside of our model's end-to-end performance, we note that the presence of any intermediate representation is useful since it makes the behavior of the system as a whole more interpretable, which is crucial in ensuring correctness. Moreover, learning a language-to-formalism mapping allows us to generalize to different execution environments, whereas a language-to-action model is only useful in the specific execution context for which it was trained.

\end{document}